\documentclass{mva_style}
\usepackage{graphicx}
\usepackage{amsmath}
\usepackage{mathrsfs}
\usepackage{algorithm}
\usepackage{algorithmicx}
\usepackage{verbatim}
\usepackage{algpseudocode}
\finalcopy 

\begin{document}
\title{Auto-Retoucher(ART) --- A Framework for Background Replacement and Foreground Adjustment}

\author{
  Yunxuan Xiao, Yikai Li, Yuwei Wu, Lizhen Zhu\\
  Shanghai Jiao Tong University\\
  800 Dongchuan RD. Shanghai, China\\
  {\tt {$\{$xiaoyunxuan, li.yi.kai ,will8821 ,zhulz98$\}$} @sjtu.edu.cn}\\
}

\maketitle

\section*{\centering Abstract}
\textit{
    Replacing the background and simultaneously adjusting foreground objects is a challenging task in image editing. Current techniques for generating such images are heavily relied on user interactions with image editing softwares, which is a tedious job for professional retouchers. Some exciting progress on image editing has been made to ease their workload. However, few models focused on guarantee the semantic consistency between the foreground and background. To solve this problem, we propose a   framework \textbf{--} ART(Auto-Retoucher)，to generate images with sufficient semantic and spatial consistency from a given image. Inputs are first processed by semantic matting and scene parsing modules, then a multi-task verifier model will give two confidence scores for the current matching and foreground location. We demonstrate that our jointly optimized verifier model successfully guides the foreground adjustment and improves the global visual consistency.
}

\section{Introduction}
\vspace{-0.2cm}
The goal of most image editing tasks is to come up with an automatic model which could replace the human labor. Consider when someone wants to turn his life photos into tourist photos in Hawaii. This would be a mission impossible for a Photoshop layman. How to replace the background and adjust the foreground location simultaneously so that the retouched image looks consistent and semantically reasonable? Now we call this specific task as \textit{Auto-Retouching} and propose a Auto-Retoucher(ART) framework to solve it.

Appearance and location between foreground and background regions are vital for the realism of composite images. Recently, the Generative Adversarial Networks successfully transfer images from one domain to another. Johnson et al. \cite{[2]} proposed a method for generating images from scene graphs. Although these generative models could process images end-to-end, the generated images mostly have poor visual quality and low fidelity in appearance. ST-GAN \cite{[1]} seeks image realism by operating in the geometric warp parameter space to find the best location, while the appearance consistency is relatively poor.

Other works in Image Harmonization focus on adjusting the appearances of foreground and background to generate realistic composites. Tsai et al. \cite{[3]} proposed an end-to-end deep convolutional neural network for image harmonization, which can capture both the context and semantic information of the composite images during harmonization. However, image harmonization models only adjust foreground appearances and retain the background region, while the location and scale of the given foreground are fixed. The edited image would not be real if the foreground objects were put in the wrong location at the very beginning. Zhao et al. \cite{[4]} present a new image search technique. Given a specific rectangle bounding box on background image, the method returns the most compatible foreground objects to paste from given candidates. However, the model is still unable to find the best location and scale of the foreground, since the bounding box is directly given by human. 

To overcome these shortcomings, we proposed Auto-Retoucher(ART) framework to generate composite images that are  harmonious both in appearance and location. A multitask verifier model will utilize semantic parsing and content features to calculate scores for both the global consistency and the spatial consistency. Then a gradient-based adjustment algorithm based on the surface of our verifier model will adjust the foreground object into the most suitable location. Since no previous work has been done on this specific task, existing datasets is not available. Therefore, we created a new dataset with 300K images for the Auto-Retouching task. Experiments are conducted on this dataset to empirically evaluate the effectiveness of our method. We prove that our model performs well on this dataset.

To summarize, the main contributions of our work are four-fold:
\begin{enumerate}
    \vspace{-0.2cm}
    \item We proposed a novel multitask verifier model to evaluate the semantic and spatial consistency, which could jointly fit both the global semantic consistency and spatial rationality.
    \vspace{-0.2cm}
    \item We introduce a gradient-based adjustment algorithm to adjust the foreground objects into a plausible location and scale.
    \vspace{-0.2cm}
    \item We construct the ART framework to solve the auto-retouching task. 
    \vspace{-0.2cm}
    \item A large scale and high quality dataset for the auto-retouching task is created.
\end{enumerate}
\section{Related Work}
\begin{description}
\item[Image Harmonization]
Traditional methods for image harmonization use color and tone matching techniques to ensure consistent appearances, such as transferring global statistics \cite{[13]}, \cite{[14]}, applying gradient domain methods \cite{[15]}, matching multi-scale statistics \cite{[16]}. Tsai. et al \cite{[3]} propose an end-to-end deep convolutional neural network for image harmonization, which can capture both the context and semantic information of the composite images during harmonization. These image harmonization methods only learn to adjust the color tone of foreground objects, but do not consider where to put the foreground images, which could be semantically inconsistent. With the help of our framework, composite images with much semantic and spatial consistency will be generated for later harmonization, further improving the quality of edited images.

In ST-GAN \cite{[1]}, spatial transformer networks are used to find geometric corrections to foreground images such that the composite images are natural and realistic. However, the model cannot ensure the color-tone consistency between foreground and background, thus selecting coherent foreground objects and background scenes still requires human interactions. In our work, after given the foreground image, no human intervention is required. The best matching background and the best location and scale will be given by a model pretrained on large dataset.

\item[Multi-task learning in CV]
 Deep multitask learning \cite{[6]},\cite{[7]} has been widely used in various computer vision problems. As Xu et al. \cite{[5]} summarized, there has been progress in many exciting tasks, such as joint inference scene geometric and semantic \cite{[8]}, face attribute estimation \cite{[9]}, contour detection and semantic segmentation \cite{[10]} . Yao et al. \cite{[11]} proposed an approach for joint learning three tasks i.e. object detection, scene classification and semantic segmentation. Hariharan et al. \cite{[12]} proposed to simultaneously learn object detection and semantic segmentation based on the R-CNN framework. We introduce the multitask learning method in our verifier model, enabling the model to jointly fit both the global semantic consistency and spatial rationality.
\end{description}

\section{Auto-Retouching Dataset}
Since the auto-retouching task is newly defined in our paper, the first challenge is the lack of data. To solve this problem, we create a large scale auto-retouching dataset. The foregrounds in this dataset are persons in different clothes. The backgrounds contain 16 different types of scenes (beach, office, desert, etc.), which fully meets our requirement for scene diversity.

The source images are from The Celebrity in Places dataset \cite{[17]}. This dataset contains 36K images of celebrities in different types of scenes, in which 4611 celebrities and 16 places are involved. We processed these images and divided all the data in our dataset into 3 categories: the positive cases, the content-level negative cases and the spatial negative cases. 

The foreground persons are first detected and cut out. Then the blank on the remained backgrounds are filled out by a content-aware filling algorithm. The filled backgrounds are later fed into a Cascade-DilatedNet \cite{[18]} to generate scene parsing maps. We denote the  foreground, background, and semantic parsing map of image $i$ as ${F_i}, {B_i}$ and ${S_i}$.

\begin{figure}[htbp]
    \centering
    \includegraphics[width=3in]{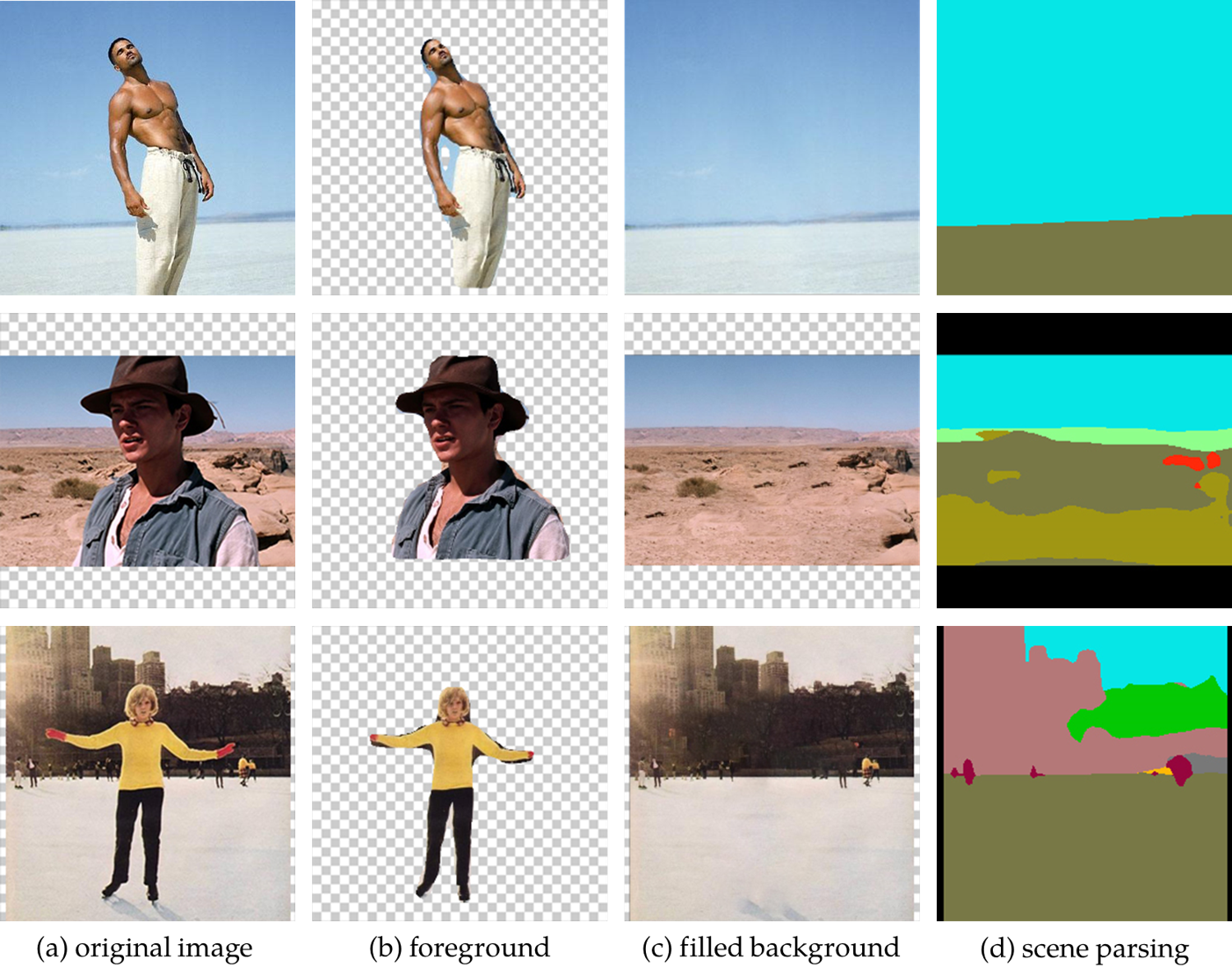}
    \caption{Samples from our dataset.}
    \label{fig:my_label}
\end{figure}

\begin{figure*}[htbp]
\centering
\includegraphics[width=6in]{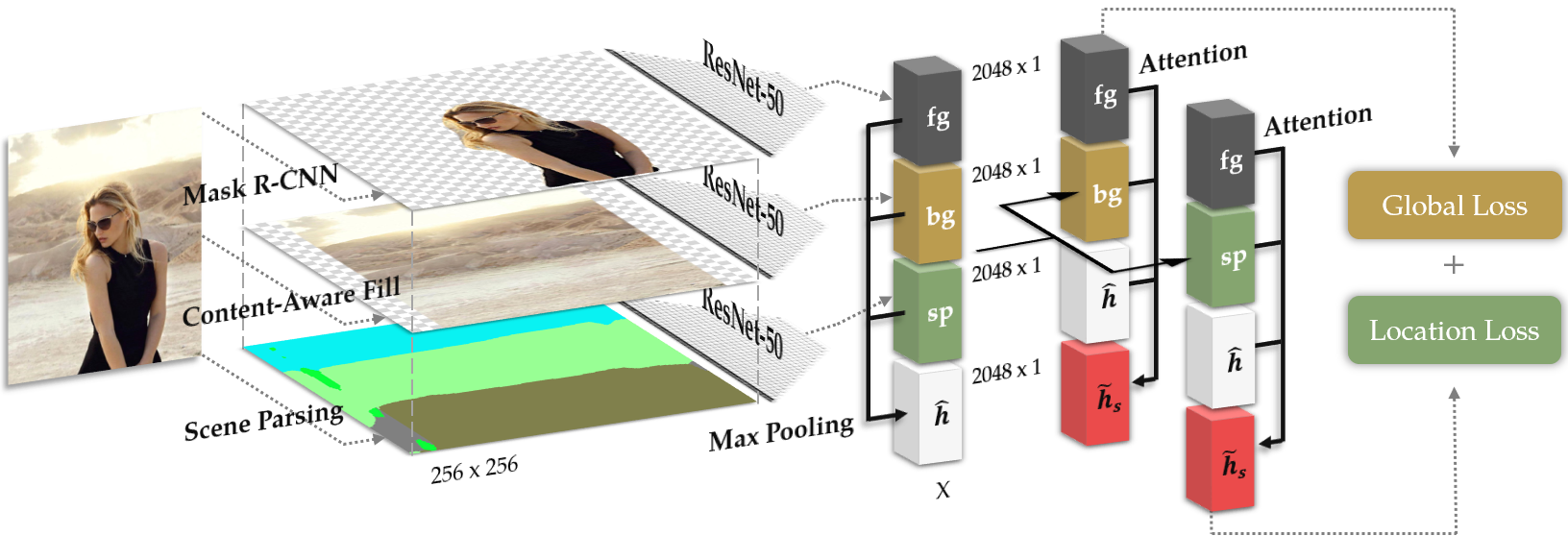}
\caption{Verifier Model Architecture}
\end{figure*}

\subsection*{Training Data Preparation and Evaluation}
To generate positive and negative cases, we hypothesize that the foreground and background of the original image are consistent both in content, regardless of the foreground's location and scale, while other fore-background combinations must be inconsistent.

Under this assumption, the positive cases are simply a set of tuples $D^+ = \{(F_i, B_i, S_i)\}$, whose content labels are True. Similarly, $D_c^- = \{(F_j, B_i, S_i)|\ i \neq j\}$ is the set of content-level negative cases, whose content labels are False. 

To create the spatial negative samples, we randomly select a series of locations and scales for the foreground objects. We assume that the randomly scattered, different scaled fore-ground patches are not consistent with the original background. Now each foreground has several substitutes $F_i^{'}$ with wrong location and scale. The spatial negative samples are $D_s^- = \{(F_i^{'}, B_i, S_i)\}$, whose content label is True based on our hypothesis. 

To evaluate the inconsistency of spatial negative cases, we proposed a spatial rationality score.
\begin{equation}
    spatial\ score = \frac{sigmoid(a - b(x/x_{max}))}{max\{r, 1/r\}} \in (0, 1)  
\end{equation}
where $x$ is the moving distance of $F_i^{'}$, $x_{max}$ is the maximum moving distance bounded by canvas and $r$ is the scaling ratio. $a=10$, $b=20$ are two constants. According to this formula, a lower spatial score indicates larger deviation from the original foreground. 

Some statistics of our dataset are listed in Table 1 below.

\vspace{-0.4cm}
\begin{table}[h] 
\caption{Statistics of training data.}
\vspace{-0.4cm}
  \begin{center}
    \begin{tabular}{c | c c c}
      \hline
      \hline
      \makebox[10mm]{} & \makebox[20mm]{content label} & 
      \makebox[20mm]{spatial score} & 
      \makebox[20mm]{number of cases}\\
      \hline
      $D^{+}$ & True & 1 & 37395   \\
      $D^{-}_s$ & True   & 0 & 74790\\
      $D^{-}_c$ & False   &  (0, 1) & 74790\\
      \hline
      \hline 
    \end{tabular}
    \label{sample-table}
  \end{center}
\end{table}
We choose accuracy as the metric of content label classification, and RMSE as the metric of spatial score regression.

\section{Auto-Retoucher Framework}
\subsection{Pipeline}
~\\

\begin{figure}[htbp]
    \centering
    \includegraphics[width=3in]{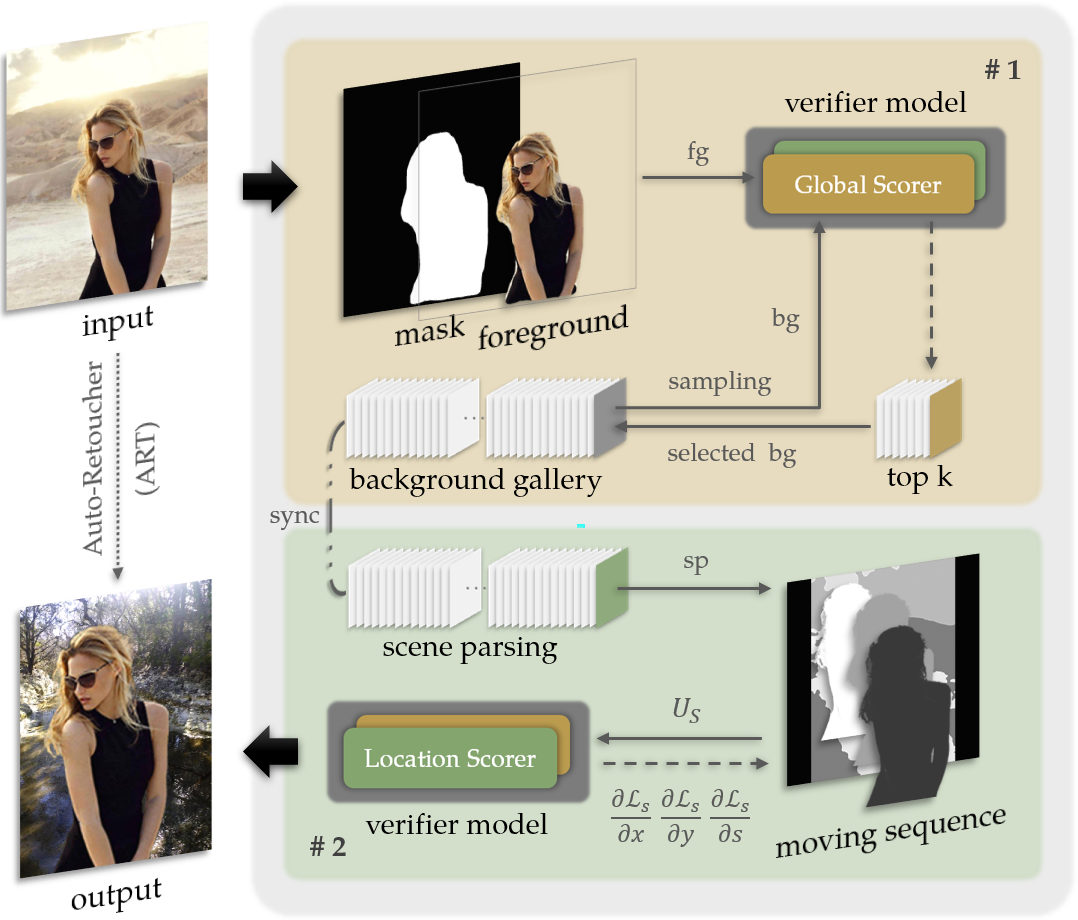}
    \caption{Framework Pipeline}
    \label{fig:my_label}
\end{figure}

\vspace{-0.5cm}
The ART framework consists of 2 stages: background selection stage and foreground adjusting stage.

In the background selection stage, several background images will be sampled from gallery. A global consistency score is given by our pre-trained multitask model. Top-k backgrounds with high global consistency scores will be sent into the next stage.

In the foreground adjusting stage, the multitask verifier model generates a score of the current foreground location and scale. The foreground images will move to a reasonable location in 3-dimensional space $(x,y,scale)$ guided by numerical gradient. Eventually, it would be placed in the plausible location with a plausible scale. This adjustment procedure will be performed multiple times with random initial points, and the best result will be selected as output.

\subsection{Multi-task Verifier Model}

Formally, we represent the tasks of our verifier model as follows: given a set of tuples $(F, B, S, y_c, y_s)$, where $F$ and $B$ are foregrounds and backgrounds, $S$ is the corresponding scene parsing representations of $B$. $y_c$ and $y_s$ are content label and spatial consistency score. The verifier model has to handle a classification task and a regression task simultaneously. The classification task is to estimate the probability $P(y_c|F,B,S)$ for global consistency. The regression task is to fit the spatial score. Our model will jointly optimize these two tasks during training.

The architecture of our model is illustrated in Figure 2. The verifier model consists of three layers: Encoding Layer, Fusion Layer and Prediction layer. 
\subsubsection{Encoding Layer: }
In our work, we utilize pretrained ResNet-50 \cite{[19]} as our encoder. These networks will be finetuned on our auto-retouching dataset. Now we denote the extracted features of foreground, background and scene parsing as $H^F$, $H^B$, $H^S$.

\subsubsection{Fusion Layer: }
Using deep features from front-end encoder, we perform a max-pooling operation to generate a sharing feature matrix $\Hat{H}$. 
$$
\Hat{H} = MaxPool([H^F, H^B, H^S])
$$

Then, the feature matrix $H^F$, $H^B$, $H^S$, $\hat{H}$ are flattened into $N\times 8192$ dimensional vectors $h_F, h_B, h_S, \Hat{h}$. Since estimating semantic consistency and spatial consistency are highly correlated, we perform a bi-attention mechanism on the deep features.
$$
\Tilde{h}_C = GELU([h^F, h^B] W_{c1} \otimes \Hat{h} W_{c2}) $$
$$
\Tilde{h}_S = GELU([h^F, h^S] W_{s1}\otimes \Hat{h} W_{s2}) $$
$$
GELU(x) = 0.5x(1+tanh[\sqrt{2/\pi}(x + 0.0447x^3)])
$$

Where $W_{c1}$,$W_{c2}$,$W_{s1}$,$W_{s2}$ project feature vectors into 30 dims. Now fusion layer generates the final representation by soft-parameter sharing:

$$
\begin{aligned}
    U_C &= [h_F, h_S, \Tilde{h}_C, \Hat{h}]\\
    U_S &= [h_F, h_B, \Tilde{h}_S, \Hat{h}]\\
\end{aligned}
$$
\subsubsection{Prediction Layer: }

The prediction layer receives fused information $L_P$ and $L_S$ and handles two prediction tasks: (1) global consistency classification (2) spatial consistency regression. Finally, fully connect layers summarize the fused vectors, and utilize the softmax and sigmoid function to produce a conditional probability $y_c$ for global consistency and a confidence score $y_s$ for spatial consistency.
$$
\begin{aligned}
     y^c &= softmax(W_1U_C + b_1) \\
     y^s &= sigmoid (W_2U_S + b_2)
\end{aligned}
$$
where $W_1, W_2, b_1, b_2$ are trainable weight matrices.

We define two independent loss for the two tasks. The content-level loss $\mathcal{L}_C$ and spatial-level loss $\mathcal{L}_S$.
$$
\begin{aligned}
\mathcal{L}_C &= -(\delta \cdot \log y^c + (1 - \delta) \cdot (\log ( 1- y^c)))\\
\mathcal{L}_S &= -\frac{1}{N}\sum_{i=1}^N(y^s_i - y_i)^2\\
\end{aligned}
$$
where $\delta \in \{1, 0\}$ is the indicator of the content label.

Finally, we combine the two losses as follow. $\lambda$ is a weight constant.
$$
\mathcal{L} = \lambda\mathcal{L_C} + (1 - \lambda)\mathcal{L_S}
$$

\subsection{Gradient Based Foreground Adjustment}
Finding the best location and scale for the foreground objects is always a tough problem in image composition tasks. In our works, we abandon the traditional method which directly regresses the optimal location and scale of the foreground. Since the performance of such methods are relatively poor due to the extremely large solution space. Instead, we hope to learn something from the spirit of Generative Adversarial Networks, using gradients of the verifier model to guide the adjustment of our foregrounds.

Here we propose a gradient-based adjustment algorithm to solve this problem: We train our verifier model to score the foregrounds that are randomly moved and scaled, where the score is defined as equation (1). Consequently, the surface of the verifier model is meaningful in the sense of gradient, as foregrounds with right location and scale get high scores and vice versa. By calculating the numerical partial gradient of location and scale $\frac{\partial{\mathcal{L_S}}}{\partial{x}} \ \frac{\partial{\mathcal{L_S}}}{\partial{y}} \  \frac{\partial{\mathcal{L_S}}}{\partial{s}}$, we can move the foreground on the surface of the verifier model by gradient ascent and finally find the best location and scale.

\begin{figure}[h]
    \centering
    \includegraphics[width=2.5in]{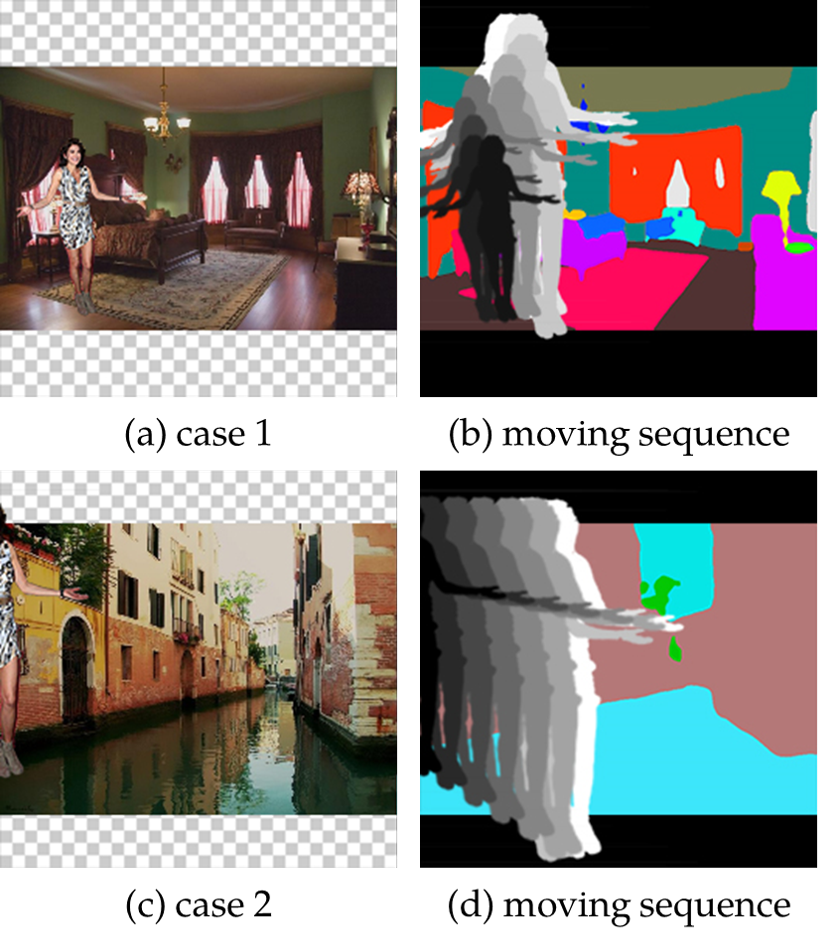}
    \caption{Visualized Results}
    \label{fig:my_label}
\end{figure}

Figure 4 shows two visualized moving sequences during the gradient ascent. In case 1, the foreground is gradually adjusted to the appropriate location and scale, proving that the surface of our verifier model is well-trained and meaningful. In case 2, the river background is globally inconsistent with foreground, thus the foreground is shifted out of the background, which is intuitively reasonable and illustrates the capability of our model to utilize the semantic information of background. 

\section{Experiments}

Our model is trained on our auto-retouching dataset, where $20\%$ of the dataset are split as test set. We use the Adam optimizer with a learning rate of 1e-5, and a dropout of 0.3 is applied for preventing overfitting. The hyper-parameter $\lambda = 0.5$ is selected by grid search. The batch size is 20 and the input image size is $256\times 256\times 3$.  We choose accuracy as the metric for the classification task and Root Mean Square Error(RMSE) for the regression task.

We did ablation experiments on our dataset to validate the performance of the verifier model and the function of the attention mechanism.

\vspace{-0.2cm}
\begin{table}[h] 
\caption{Performance on the test set}
  \begin{center}
    \begin{tabular}{c | c c}
      \hline
      \hline
      \makebox[10mm]{} & \makebox[20mm]{Task1 Accuracy} & 
      \makebox[20mm]{Task2 RMSE}\\
      \hline
      base model & $87.5\%$ &   $0.354$ \\
      -Attention &  $86.6\%$  & $0.359$ \\
      \hline
      \hline 
    \end{tabular}
    \label{sample-table}
  \end{center}
\end{table}

The result in Table 2 shows that our model does well on the background selection task and foreground adjustment task. Also, the attention mechanism actually improves the overall performance. 




\section{Conclusion}
We focused on a specific auto-retouching task and designed a novel framework for this task. The ART framework is able to replace the image background and adjust the foreground location and scale simultaneously, while keep the edited image semantically and visually harmonious. We first introduce multitask learning method to combine two auxiliary losses to help the verifier model concentrate on content-level consistency and spatial consistency respectively, and then utilize the confidence scores to guide the background selection and foreground adjustment. We created an auto-retouching dataset containing 300K images. Our system has achieved good visual performance on human judgment. Looking forward, we plan to design new structures for verifier model to handle more complicated circumstances.

\end{document}